# Hybrid Metaheuristic Vehicle Routing Problem for Security Dispatch Operations

Nguyen Gia Hien Vu[1], Yifan Tang[1], Rey Lim[2], and G. Gary Wang[1,3]

**ABSTRACT**

This paper investigates the optimization of the Vehicle Routing Problem for Security Dispatch (VRPSD). VRPSD focuses on security and patrolling applications which involve challenging constraints including precise timing and strict time windows. We propose three algorithms based on different metaheuristics, which are Adaptive Large Neighborhood Search (ALNS), Tabu Search (TS), and Threshold Accepting (TA). The first algorithm combines single-phase ALNS with TA, the second employs a multiphase ALNS with TA, and the third integrates multiphase ALNS, TS, and TA. Experiments are conducted on an instance comprising 251 customer requests. The results demonstrate that the third algorithm, the hybrid multiphase ALNS-TS-TA algorithm, delivers the best performance. This approach simultaneously leverages the large-area search capabilities of ALNS for exploration and effectively escapes local optima when the multiphase ALNS is coupled with TS and TA. Furthermore, in our experiments, the hybrid multiphase ALNS-TS-TA algorithm is the only one that shows potential for improving results with increased computation time across all attempts.

1. **INTRODUCTION**

Since its first introduction (Dantzig & Ramser, 1959), the Vehicle Routing Problem (VRP) has remained an active research topic in combinatorial optimization due to its multi-attribute nature and wide-ranging applications (Erdoğan, 2017; Vidal et al., 2013). As highlighted by Raza et al. (2022), the main objective of VRP is to determine the route that allows different customer requests to be efficiently serviced using a fleet of vehicles, costs such as that of time or travelling distance minimized, and different problem requirements fulfilled. Normally, each customer is serviced just once with only one vehicle. Each vehicle often has a limited capacity, and all vehicles must start and end at a depot (Raza et al., 2022).

Throughout the years, efforts have been made to keep researching VRP variants which are defined when constraints are added to problems, subject to the goods, service, customers, and fleets in concern (Baldacci et al., 2012; Elshaer & Awad, 2020; Kallehauge, 2008; Shahbazian et al., 2024). Consequently, VRP has evolved to include numerous variants (Sarbijan & Behnamian, 2023) to reflect real-world complexities and diverse operational constraints. To illustrate, just for basic VRP variants, Elshaer & Awad (2020) have mentioned more than ten VRP categories, including the following three main ones:

- Vehicle Routing Problem with Time Windows (VRPTW): each vehicle must deliver goods to customers within specified time spans.
- Split-Delivery Vehicle Routing Problem (SDVRP): customers' demands can be split and delivered by multiple vehicles on different routes.

---

[1] Product Design and Optimization Laboratory, Simon Fraser University, Surrey, BC, Canada
[2] Simon Fraser University Alumnus, BC, Canada
[3] Corresponding author: Email gary_wang@sfu.ca Tel: 778 782 8495

- Open Vehicle Routing Problem (OVRP): vehicles are not required to return to the original depot after their delivery services are completed.

As an example, VRPTW conceptually means that each customer must be serviced during a certain time interval, known as a time window (Baldacci et al., 2012; Elshaer & Awad, 2020). If the vehicle arrives at the customer earlier than the time window, it must wait until the time window starts. However, if the vehicle arrives after the deadline of the time window, it violates the problem constraints (Kallehauge, 2008). In practice, multiple other constraints can also be added to VRPTW to create different VRPTW variants, such as the existence of different depots (Milburn, 2012). In this trend, recent years have also observed the introduction of new VRP variants, such as Green-VRP (Asghari & Al-E-Hashem, 2021), VRP with drones (Viloria et al., 2021), or electric VRP (EVRP, Kucukoglu et al., 2021).

With such numerous variants, VRP can be broadly applied in many fields (Raza et al., 2022) whether they are commercial, civil, or military (Vidal et al., 2013). However, even in such a family of VRP variants, to the best of the authors' knowledge, VRP for Security Dispatch (referred to as VRPSD hereinafter) receives little attention. Semantically, security dispatching refers to sending people, vehicles or any resources as required to nominated locations for security purposes, for example, to protect targets such as persons, buildings, or entities including businesses or government agencies (Press, 2004). As emphasized by DePasquale (1998), while security operations are to prevent threats or breaches from happening proactively or to respond to emerging incidents reactively, resources must be optimized for key priorities

In recent years, the demand for private security has risen worldwide (Blackstone et al., 2023), implying the growing need for security dispatch. Given the objective and nature of security operations, customers have become increasingly discerning and challenging with their highly variable and complex requirements (Mancini, 2018), such as in security and patrolling tasks. In other words, such services should be available in myriads of customized and exclusive modes and specifications beyond time-window requirements. In any case, they should simultaneously meet numerous industry-specific requirements, particularly in terms of:

- Optimizing timing, routing, and costs (Raza et al., 2022),
- Adhering to operational conditions (Raza et al., 2022), and
- Assuring routing unpredictability to reduce risks of bad actors guessing and trying to fail dispatch missions (Mancini, 2018).

Considering such requirements, a number of constraints must arguably be satisfied in VRPSD. For example, security dispatch assignments can entail single or multiple visits each night. Noticeably, each visit to a specific location or site might require the same or different types of constraints, such as different time windows or hardware requirements. Also, each customer often desires different amount of service time for different targets. For instance, an outside patrol may normally be shorter than a full interior patrol. For a specific site, customers may need the same service multiple times a day, with each request varying in the amount of time required, even in the absence of unforeseen issues. Additionally, for each site, the same service must not be provided back-to-back, i.e., there must be a gap between consecutive services for the same site. However, when customers require different types of services for the same site, providing services one after the other is not be considered as back-to-back violations. In addition, certain dispatch tasks demand peculiar constraints including highly precise timing, such as site lock-up or escorts, with time windows as narrow as 10 minutes. Moreover, customers might also need more than one routing solution to enhance unpredictability. Notably, a number of these constraints can be found in the Surveillance Patrol Vehicle Routing Problem described by Mancini (2018). Analytically, no matter what type of service is demanded, routing problems and precise time windows are always involved as core constraints.

Such widely diversified and unique requirements in the security dispatch industry are reflected in the request made by the industrial partner, who provides the dataset as the background for this paper. Specifically, the requirement is to optimize:

- The number of vehicles and drivers, or shifts, that are to be assigned to the most suitable routes that satisfy a set of time and service constraints. In this paper, one shift is equivalent to one full route, and therefore they can be used interchangeably, and
- Routes to minimize the total travel time to different sites.

To make it clearer, the goal herein is to optimize the timing and routing process in dispatch assignments for security and patrolling purposes. This necessitates the development of an automated algorithm for optimal solutions as opposed to manual processes. Expectedly, an effective algorithm should identify potential solutions that can reduce costs, increase revenue, and improve competitiveness (Mancini, 2018). Ideally, such timing requirements and routing optimization tasks have the potential to imply considerable time-window-shortening applications not only in security operations but also in dispatch-related sectors. Examples can include the delivery industry or medical services, where precise timing and strict time windows can help enhance customer satisfaction.

In this line, this paper aims to develop an algorithm to automatically address VRPSD for timing and routing optimization in security dispatch, utilizing an actual input patrolling dataset provided by the industrial partner. Notably, VRPSD shares characteristics of VRPTW due to the time window requirements, yet with additional security dispatch constraints. The proposed approach integrates three different metaheuristics, namely Adaptive Large Neighborhood Search (ALNS), Tabu Search (TS), and Threshold Accepting (TA).

Within such a scope, the remainder of this paper is structured as follows: Section 2 provides a brief review of VRP solving methods and then presents the objectives and requirements for this paper as a problem definition which is to be addressed with VRPSD. Section 3 discusses our experimental setup, including the detailed implementation of the three individual metaheuristics. Section 4 presents the main algorithms as the targets of this paper, i.e., how different metaheuristics can be combined together. Section 5 analyses and discusses results of different experiments. Finally, Section 6 summarizes the findings and recommends possible research in the future.

## 2. LITERATURE REVIEW AND PROBLEM DEFINITION

### 2.1 VRP Solving Methods

As discussed in Section 1, there exist considerably numerous VRP variants, each of which has its unique requirements and characteristics. This truly challenges any effort to find a universal and up-to-date VRP solver (Erdoğan, 2017). Additionally, solving VRP is computationally challenging due to its NP-hard nature, resulting in different approaches created for the last six decades (Vidal et al, 2013). For example, exact methods, including the branch-and-cut and set partitioning algorithms (Baldacci et al., 2012), are effective for small instances, but they often have scalability issues for larger ones (Raza et al., 2022; Vidal et al., 2013). For instance, most VRPTW problems reviewed by Baldacci et al. (2012) have approximately 100 or less customers, which can be smaller than the size and complexity of real-world VRP instances (Erdoğan, 2017). When the number of customers goes beyond 100, depending on the variants, VRP exact method might struggle to find a solution in a reasonable amount of time (Kumar & Panneerselvam, 2012).

To overcome such drawbacks, specific heuristic and metaheuristic approaches have been widely adopted (Vidal et al, 2013). Heuristic algorithms, such as saving methods or sweep algorithms (Vidal et al, 2013), are designed for particular problems.

However, they are often problem-dependent (Talbi, 2009) or might get stuck in the first encountered local optimum (Vidal et al, 2013).

Among all, metaheuristic algorithms stand out because of their capabilities of exploring the search space efficiently for different computationally expensive problems (Talbi, 2009). Multiple metaheuristics, such as Ant Colony Optimization (ACO), Guided Local Search (GLS), or Genetic Algorithm (GA), have been applied to achieve decent routing solutions (Elshaer & Awad, 2020). Noticeably, current metaheuristics are seemingly not designed specifically to deal with constraints inherent in VRPSD. Firstly, the current algorithms possibly do not consider the need to visit one customer multiple times. The review paper by Elshaer and Awad (2020) mentions that all customers should be visited exactly once, by exactly one route. Secondly, the algorithms seemingly lack focus on introducing extra randomness to create considerably different routing solutions which can be particularly meaningful for enhancing unpredictability. For example, Labdiad et al. (2021) uses the roulette wheel as the only source of randomness, while ALNS operations are deterministic. Thirdly, many current metaheuristics might not be developed to deeply and specifically emphasize VRPSD requirements and objectives, such as strict time window constraints and shift minimization. As exampled by Labdiad et al. (2021), there is no operator in the algorithm that repairs a route for late-serviced customers. Similarly, Mancini (2018) uses the Greedy Randomized Adaptive Search (GRASP) algorithm to tackle VRP in the Surveillance domain which shares a number of similar requirements with VRPSD as referred to in Section 1. However, Mancini (2018) applies soft time window constraints, meaning that solutions that violate these time windows are not deemed infeasible. As a result, this approach does not have effective or direct methods for addressing time window violations. Additionally, Mancini (2018) assumes that the problem instance operates within an 8-hour time frame and does not set a limit on the number of shifts when creating the initial solution. This approach could potentially hinder the objective of optimizing the number of shifts as proposed in Section 1. Arguably, these existing approaches cannot be applied to solve VRPSD without significant modifications.

Additionally, the integration of emerging technologies has recently expanded the scope and applicability of VRP, such as the use of Artificial Intelligence (AI) and Machine Learning (ML) (Shahbazian et al., 2024). However, these approaches reveal certain problems, such as expensive label data, error transmissions, or inappropriate model selection (Wang, 2022).

## 2.2 Problem Definition

This subsection presents the problem definition of VRPSD to be dealt with in this paper. Though this definition is formulated as general as possible to capture common features of a VRPSD, the authors acknowledge that this definition does not attempt to entail all possible constraints in a real-world VRPSD instance. All the constants used in the problem definition can be adjusted to reflect the actual situation.

"Let $G = (V, A)$ be a network where $V$ is the set of vertices and $A$ is the set of arcs. Each vertex in $V$ represents either a customer $c_i$ or the depot $d$. Each customer can request single or multiple different types of services or request the same service but with different time windows. Each customer cannot be serviced back-to-back, i.e., there must be a gap between consecutive services for the same site.

Each arc in $A$ represents the shortest time to travel between two vertices $i$ and $j$, denoted as $t_{ij}$, which remains constant throughout the solving process. A known set of vehicles is available, all starting and ending their shifts at the depot $d$. Each vehicle is operated by a driver, and collectively, these drivers must service all customers $c_i$. Servicing a specific customer requires a specific duration $q_i$.

The service at customer $i$ can begin no earlier than time $a_i$ and must be completed by time $b_i$. A driver may arrive at the site before $a_i$, but cannot start the service until $a_i$. The time the driver spends waiting for the service window to start is called the downtime. However, if the service extends beyond $b_i$, the time window constraints are violated. Drivers are not permitted to provide the same service at the same location consecutively.

Additionally, each driver's shift is subject to the following constraints:

1. The shift must not exceed 12 hours; otherwise, it violates the constraints.
2. Although shifts shorter than 8 hours are acceptable, they are not preferred.
3. Each shift includes 15 minutes for check-in and 15 minutes for check-out.
4. Drivers must take at least a 30-minute break during each shift, but the timing of the break is flexible.
5. The shift duration, if possible, is preferably to be similar to each other.

The objective herein is to determine the set of routes, or shifts, that minimizes either the total cost, or the total time across all shifts."

While inherently featuring many characteristics of the traditional VRPTW, VRPSD is designed to emphasize distinct industry-specific requirements as discussed in Section 1, with particular focus on:

- Satisfying narrow time window requirements, especially when special services might demand highly precise timing,
- Having each site visited once or multiple times each day, and
- Avoiding same sites to be serviced back-to-back if multiple visits are required.

3. **METHODOLOGY**

Considering features of metaheuristic algorithms and distinct demands in security and patrolling services, this section presents the methodology to solve the VRPSD problem definition as stated in Subsection 2.2.

3.1 General Considerations

In this study, the VRPSD is optimized for a 24-hour cycle. Notably, in line with the constraints stated in Subsection 2.2, the solution should be considered infeasible if one of the following conditions occurs:

- A customer is serviced back-to-back.
- One shift is longer than the limit (12 hours).
- The service deadline for a customer is violated.

Also, the number of shifts and the corresponding start time is determined in advance and stays constant throughout the optimization process. However, if a shift starts from and ends at the depot immediately, such a shift is considered redundant and is eliminated from the final output.

Furthermore, to deal with multiple requests from the same customer, if a location needs to be visited multiple times in a day, each visit is treated as a distinct customer. With this modification, each customer is now serviced exactly once.

3.2 Initial Solution

The initial solution starts with a list of empty routes. To begin with, the list of unassigned customer requests includes all customers inside the 24-hour cycle. This list is sorted in the order of deadlines, i.e., the customers with the earliest and latest deadlines are at the top and end of the list, respectively. For each iteration, the customer at the top of the list is chosen to be inserted into the current solution and then removed from the list of unassigned customers. Markedly, this customer is inserted into the solution with the greedy algorithm, detailed by the description shown in Pseudo-code 1. This initial solution is utilized as the input for all attempts and algorithms. The detailed approaches for different algorithms are discussed in Section 4.

| Pseudo-code 1. Initial Solution Creation |
|---|
| 1: $S \leftarrow$ Empty Solution |
| 2: $L \leftarrow$ List of All Customers in the Sorted Order |
| 3: While $L$ is not empty |
| 4:     Pick one request $i$ from the top of $L$ |
| 5:     $U \leftarrow$ List of all routes that can service customer $i$ without creating violations |
| 6:     If $U$ is empty: |
| 7:         Choose one random route $T$ from $S$ |
| 8:         Append $i$ to the end of $T$ |
| 9:         Update route $T$ in $S$ |
| 10:    Else: |
| 11:        Pick the route $T$ from $U$ that can finish servicing its current route and arrives at $i$ the earliest |
|           // If multiple routes can finish their current routes and arrive at $i$ simultaneously, the first route in $U$ is selected. |
| 12:        Append $i$ to the end of $T$ |
| 13:        Update route $T$ in $S$ |
| 14:    End if |
| 15: End while |

Pseudo-code 1 iteratively assigns customer requests to routes by processing them from a sorted list (lines 1-2). For each customer, it checks for routes that can service the request without violating constraints (lines 4-5). If no valid route is found, a random route is selected, and the customer is assigned to it (lines 6-10). If feasible routes are available, Pseudo-code 1 chooses the route that can complete its current route and arrive at the customer the earliest. Once a route is selected, the customer is appended to it, and the route details are updated (lines 10-14). This process repeats until all customers are assigned to routes.

### 3.3 Objective Function

As a crucial component in solving the VRPSD optimally, the objective function defines the criteria for evaluating the quality of a given solution (Arora, 2017). Noticeably, the goal herein is to optimize the total cost, or time, associated with the routes taken by the fleet of vehicles, while minimizing the number of violations. Hence, the objective function includes the following cost components:

- Cost of shift setup: The shift setup cost refers to the fixed costs incurred for using each additional shift, provided that the additional shift serves at least one actual customer.
- Cost of time: This component captures the total time spent by all vehicles during their shifts, including the travel time between customer locations and depot, servicing time, and downtime. Notably, the extra cost is introduced for the downtime since it is inefficient and normally unwanted.

- Cost of violations: This component represents penalties incurred when constraints are violated. Violations include:
  - Missing customer deadline window.
  - Servicing customer back-to-back.
- Cost of improper shift durations: This component is introduced when the shift duration is beyond the limit of 12 hours as provided in Subsection 3.1, i.e., when the shift is too long or too short.

### 3.4 Metaheuristics

This study employs three different metaheuristic algorithms, each of which is chosen for its respective unique approach to optimization. They are Adaptive Large Neighborhood Search (ALNS), Tabu Search (TS), and Threshold Accepting (TA), which are discussed in detail in the following subsections with the emphasis on their key principles and operations.

#### 3.4.1 Adaptive Large Neighborhood Search (ALNS)

For this paper, ALNS is selected because its iterative process allows the algorithm to explore bigger regions of the search space and hence can find improved solutions (Kuyu & Vatansever, 2024). In principle, ALNS iteratively destroys and repairs portions of the existing solution (Voigt, 2025). During each iteration, ALNS selects one operation to destroy the current solution partially, then selects a separate repair operator to repair the destroyed solution, and thus creates a new solution. The probability whether a specific operation is chosen depends on its past performance, predefined initial weights, and a roulette wheel mechanism (Voigt, 2025). The new solution is accepted with the TA algorithm, which is discussed in detail in Subsection 3.4.2.

This process utilizes the existing library from Wouda and Lan (2023). This library includes and allows us to adjust many ALNS operations, including:

- Adaptive search and roulette wheel selection mechanism, and
- Adaptive weight adjustment.

In addition to such an existing library, this paper also implements specific repair and destroy operations. Notably, these operations are chosen carefully from many different operations summarized by Voigt (2025), based on which each of the destroy operations is described by two criteria:

1. Which routes, or shifts, are selected to be destroyed?
2. What criteria are used to determine which customers to be destroyed?

The selected customers are removed, or destroyed, from the solution one by one, and the order in which they are removed is recorded.

In repair operators, removed customers are iteratively inserted back to the destroyed solution one by one. Similar to destroy operators and the summarization from Voigt (2025), repair operators can also be defined with the following questions:

1. Which routes, or shifts, are selected to be inserted?
2. What criteria are used to determine the order in which the sites should be inserted?
3. What position should be inserted in the chosen shift?

Accordingly, 17 destroy operators and 17 repair operators are implemented. The detailed description of each operator is in Appendix 1 of this paper. Notably, the same destroy and repair operators are applied to all algorithms described in Section 4 (Main Algorithms).

### 3.4.2 Threshold Acceptance (TA)

Basically, this paper utilizes TA in combination with ALNS with due considerations of its advantages. In this combination, if the new solution created by ALNS is better than the existing solution, it will be accepted as a new current solution. Additionally, TA can accept worse solutions created by ALNS only if they fall within a predefined threshold, progressively narrowing this threshold to refine the solution (Tarantilis et al., 2004). Moreover, TA has simple structure, general applicability and computational effectiveness on different combinatorial optimization problem, while it can prevent the algorithm from getting stuck in local optima (Tarantilis et al., 2004). TA threshold curve is implemented as a linear line using exactly the same library from Wouda and Lan (2023). The detailed implementation is discussed in Subsection 4.2 (Algorithm 1).

### 3.4.3 Tabu Search (TS)

In principle, TS utilizes the Tabu List to prevent revisiting recent solutions, enabling a focused yet diversified exploration of the solution space (Ahmed and Yousefikhoshbakht, 2023). In this paper, the TS implementation swaps the position of two sites and thus creates a new solution. This move is then appended into the Tabu list, which utilizes a queue structure, i.e., "first in, first out" operations.

The TS operations are also selected using roulette wheel mechanisms which are implemented on the basis of the existing library from Wouda and Lan (2023). However, unlike the usage of TA (Subsection 3.4.2) in combination with ALNS (Wouda & Lan, 2023), all new TS solutions are accepted to encourage exploration. TS implementation includes the following two operations:

1. Random swap.
2. Long arc swap as described in Pseudo-code 2. In Pseudo-code 2, arcs are categorized as either long or short ones based on their travel time. In this paper, an arc, which represents the route between two stops, is considered long if neither of the stops is the depot, and the travel time between them exceeds 9.5 minutes. Conversely, an arc is classified as short if neither stop is the depot and the travel time is under 8.5 minutes. The focus of this operation is to eliminate long arcs, as these arcs can significantly increase the total travel time.

| Pseudo-code 2. Tabu Search 2$^{nd}$ operators |
| --- |
| 1: $S \leftarrow$ Existing Solution |
| 2: Pick Random Route 1 and Random Route 2 (Route 1 and Route 2 can be the same) |
| 3: In route 1, search for stop A and stop B such that AB is a long arc |
| 4: In route 2, search for stop C and stop D such that CD is a long arc |
| 5: If could not find the pair AB or could not find the pair CD: |
| 6:     Return |
| 7: If A==C or B==C or A==D or B==D: |
| 8:     Return |
| 9: *Elif* AD and BC are short arcs: |
| 10:     Switch B and D if arc BD switch not found in Tabu List |
| 11:     Append BD to Tabu List |
| 12: *Elif* AC and BD are short arcs: |
| 13:     Switch B and C if arc BC switch not found in Tabu List |
| 14:     Append BC to Tabu List |
| 15: End if |

In each iteration of Pseudo-code 2, two random routes are selected, and two distinct pairs of stops with long arcs are identified within these routes (lines 2-4). If there is no valid pair, no action is taken (lines 5-8). If valid pairs are found, Pseudo-code 2 checks whether swapping the stops improves the solution. Specifically, the pseudo-code swaps the stops only if the resulting arcs are short and the swap is not in the Tabu List. If a valid swap is made, the new arc is added to the Tabu List to avoid revisiting the same move (line 9-14).

Similar to ALNS, a new solution is generated after each TS iteration. If the solution is accepted, it is assessed to check if it qualifies as the new global solution. If it does, the global solution is then updated.

## 4. MAIN ALGORITHMS

This section describes a benchmark metaheuristic as Algorithm 0 using the Genetic Algorithm (GA), and then three different algorithms, Algorithms 1-3, which are incrementally developed in this paper. Conceptually, for Algorithm 0, each time the new offspring is evaluated, it is counted as one iteration. However, for Algorithms 1-3, one iteration is counted each time a newly found complete solution is evaluated to determine if it represents the new global solution. Each of these four algorithms is run for two hours. Additionally, Algorithm 3 (Subsection 4.4) is also tested with the run time of 10 hours to observe the time effect on its performance.

### 4.1 Algorithm 0 – Benchmark Algorithm

As this is the first study to explore VRPSD, there is no existing and open access library specifically designed to address this particular VRP variant, to the best of the authors' knowledge. Consequently, the benchmark algorithm, Algorithm 0, is adapted from a VRP solver available on Github, which utilizes the Steady-State GA (SSGA) (Yu, 2021; Jenkins et al., 2019). The SSGA operates by randomly selecting two parents, generating one offspring, and replacing the least fit individual in the population with the new offspring (Jenkins et al., 2019). SSGA is beneficial since it performs only a single function evaluation for each child every cycle, which helps streamline the overall running time (Jenkins et al., 2019).

### 4.2 Algorithm 1

In Algorithm 1, based on the example code provided by Wouda and Lan (2023), ALNS and TA are applied together, and the resulting detailed algorithm is shown in Algorithm 1.

| |
|---|
| **Algorithm 1. ALNS and TA algorithm** |
| 1: $S \leftarrow$ Initial Solution |
| 2: Define operator weights, decay rate, TA algorithms parameters, and other parameters |
| 3: Select a random seed |
| 4: While the running time limit is not reached do: |
| 5:     Select the destroy and repair operators, and run the described ALNS algorithm |
| 6:     Determine whether the new solution is accepted using the TA algorithm |
| 7:     If the new solution is accepted and is the new global best |
| 8:         Update the global solution |
| 9:     End if |
| 10: End while |
| 11: Return the best solution |

Algorithm 1 iteratively improves the initial solution $S$ (Pseudo-code 1) by actively searching within a specified wall-clock algorithm running time limit of two hours, using ALNS. The running time should be sufficiently long so that the total travel time experiences no reduction in at least 60% of the attempts in the final half of the total iterations. The algorithm starts by defining necessary parameters (lines 2-3). In each iteration, the destroy and repair operators are chosen and ALNS is executed (line 5). The new solution is then evaluated using TA to determine whether it should be accepted (line 6). If the new solution is accepted and better than the current global solution, the global solution is updated (lines 7-8). This process continues until the running time limit is reached (line 4), and the best solution found is returned (line 11).

As documented by Wouda (2024) and Wouda and Lan (2023), the initial weights for all operators are set to 1. For each iteration, the new candidate can fall into one of the following categories:

1. New global solutions
2. Better solutions, but not the global best
3. Accepted
4. Rejected

Corresponding to these four categories are the following four respective scores $s_j$: [6, 5, 1, 0]. Additionally, a "decay rate" of 0.9 is applied throughout the process. After each operation, the updated weight $w_{new}$ can then be calculated using the old weight $w_{old}$, decay rate $d$, and score $s_j$ as (Wouda, 2024; Wouda & Lan, 2023):

$$w_{new} = d * w_{old} + (1 - d) * s_j \qquad \text{Equation 1}$$

The TA acceptance criteria is implemented using the $RecordtoRecordTravelFunction.autofit$ function by Wouda (2024) and Wouda and Lan (2023). Accordingly, this function generates a linear line as the threshold for the TA, with two input points serving as the starting and ending points of the line. However, this TA threshold line is modified, so the starting point equals to the initial solution objectives, the end point equals to 0.02 of the starting point, and the number of iterations is 9,000.

### 4.3 Algorithm 2

Algorithm 2 is similar to Algorithm 1 since it also utilizes ALNS and TA. However, the main difference is that Algorithm 2 employs ALNS in a multiphase process as described below.

| |
|---|
| **Algorithm 2 . Multiphase ALNS and TA algorithm** |
| 1: $S \leftarrow$ Initial Solution |
| 2: While the running time limit is not reached do: |
| 3:     Define operator weights, decay rate, TA algorithms parameters, and other parameters |
| 4:     Select a random seed |
| 5:     Set the time limit for one ALNS phase |
| 6:     While the time limit for one ALNS phase is not reached do: |
| 7:         Select the destroy and repair operators, and run the described ALNS algorithm |
| 8:         Determine whether the new solution is accepted using the TA algorithm |
| 9:         If the new solution is accepted and is the new global best |
| 10:            Update the global solution |
| 11:         End if |
| 12:     End while |
| 13: End while |
| 14: Return the best solution |

Similar to Algorithm 1, Algorithm 2 starts with an initial solution $S$ using Pseudo-code 1 (lines 1-2). However, for each ALNS phase, the new parameters and phase running time limit is determined (lines 3-5). Within each ALNS phase, the destroy and repair operators are selected, and ALNS is executed (line 7). The new solution is evaluated using TA to determine whether it should be accepted (line 8). If the new solution is accepted and becomes the global best, the global solution is updated (lines 9-10). This continues until the time limit for each ALNS phase is reached (line 6). Then, such a process repeats until the overall time limit is met (line 2). Finally, the best solution found during the process is returned (line 14).

The main difference between Algorithm 1 and Algorithm 2 is that, after the time limit for each ALNS phase is reached, the new parameters (line 3-5) are reselected, and a new ALNS phase is conducted. For testing purposes, the same total running time of two hours is applied to both Algorithm 1 and Algorithm 2. However, since Algorithm 2 involves a multiphase ALNS, the total running time is split into two equal intervals for the two ALNS phases. The other hyperparameters for both ALNS and TA algorithms are similar to that of Algorithm 1. However, in the second TA phase, the TA curve is adjusted: the starting point is set to the current global solution's objectives, the endpoint is 2% of the starting point, and the number of iterations is 9,000.

## 4.4 Algorithm 3

Algorithm 3 uses the multiphase ALNS and TA, similar to Algorithm 2. However, the highlight herein is the use of TS, as shown below.

| |
|---|
| **Algorithm 3. Hybrid multiphase ALNS** |
| 1: $S \leftarrow$ Initial Solution |
| 2: While the running time limit is not reached do: |
| 3:     Define operator weights, decay rate, TA algorithms parameters, and other parameters |
| 4:     Select a random seed |
| 5:     Set the time limit for one ALNS phase |

| | |
|---|---|
| 6: | While the running time limit for one ALNS phase is not reached do: |
| 7: | Select the destroy and repair operators, and run the described ALNS algorithm |
| 8: | Determine whether the new solution is accepted using the TA algorithm |
| 9: | If the new solution is accepted and is the new global best |
| 10: | Update the global solution |
| 11: | End if |
| 12: | End while |
| 13: | Define operator weights, decay rate, TA algorithms parameters, and other parameters |
| 14: | Select a random seed |
| 15: | Select the time limit for one ALNS phase |
| 16: | While the running time limit for one ALNS phase is not reached do: |
| 17: | Select the destroy and repair operators, and run the described ALNS algorithm |
| 18: | Determine whether the new solution is accepted using the TA algorithm |
| 19: | If the new solution is accepted and is the new global best |
| 20: | Update the global solution |
| 21: | End if |
| 22: | End while |
| 23: | Define TS parameters |
| 24: | Set the time limit to run TS |
| 25: | While the time limit for TS is not reached do: |
| 26: | Select and run TS operators |
| 27: | If the new solution is accepted and is the new global best |
| 28: | Update the global solution |
| 29: | End if |
| 30: | End while |
| 31: End while | |
| 32: Return the best solution | |

Algorithm 3 iteratively improves the initial solution $S$ (Pseudo-code 1) using the following steps:

1. Key parameters are defined, including operator weights, decay rate, and settings for TA (lines 3-4). A random seed is selected (line 4), and a time limit for one ALNS phase is set (line 5).
2. Within each ALNS phase, this algorithm selects destroy and repair operators and runs ALNS (lines 7-8). If the new solution is better than the global best, it is accepted, and the global solution is updated (lines 9-10). This continues until the time limit for the ALNS phase is reached (line 6).
3. After the first ALNS phase, the parameters are redefined (lines 13-15), and the process repeats for another ALNS phase (lines 16-22).
4. Once the ALNS phases are completed, this algorithm proceeds to the TS phase (lines 23-25). The TS parameters are redefined (line 23), and a time limit for running TS is set (line 24). Within the TS phase, this algorithm selects and runs

TS operators (lines 26-27). If the new solution is better than the global best, it is accepted, and the global solution is updated (lines 28-29). This TS phase continues until the time limit for this phase is reached (line 25).

5. Once the round is completed, the entire process repeats until the overall time limit is reached (line 2), and the best solution found during the iterations is returned (line 32). Notably, the TS list is maintained for the next iteration.

The hyperparameters for both ALNS and TA in this algorithm are similar to those used in Algorithm 2. For TS, the parameters, including weights and decay rate, are also similar to ALNS, with the exception that all solutions are always accepted, even if the new solution is worse than the current one. The TS list has a fixed length of 10. For each new phase of the algorithm, TS hyperparameters are reinitialized. However, the content of the Tabu list remains fixed for the next phase. Additionally, since one TS iteration is much faster than one ALNS or TA iteration, approximately 10% of the total running time in each round is allocated to TS, with the remaining time divided between the two ALNS phases.

## 5. RESULTS AND DISCUSSION

### 5.1 Experimental Environments and Results

Four proposed algorithms (Algorithms 0, 1, 2, and 3) are evaluated in a scenario involving 251 customer requests over a 24-hour period, utilizing the actual input patrolling dataset provided by the industry partner in this study. This detailed dataset is available at: https://github.com/vuk1716/VRP-For-Security-Dispatch-Data. The simulation is conducted with a fleet of seven shifts, and the shift start times are decided based on their current operational practices. Travel times between locations are obtained using Google Maps API at 9:00 AM on Jun 25th, 2024. Tests are performed on a system equipped with an Intel i7-10750H processor and 32GB of RAM.

The cost matrix is computed to encompass both the travel time and the time required to service the next site. Table 1 presents a portion of the cost matrix to facilitate better visualization.

| Stop ID | 50006 | 43440 | 43448 | 67933 |
|---|---|---|---|---|
| **50006** | 0 | 1439 | 1481 | 2759 |
| **43440** | 4559 | 0 | 0 | 1800 |
| **43448** | 4559 | 0 | 0 | 1800 |
| **67933** | 4559 | 480 | 522 | 0 |

Table 1. A section of the cost matrix

Specifically, the value at the intersection of Row $i$ and Column $j$ in the cost matrix represents the total time needed to travel from the site in Row $i$ to the site in Column $j$, as well as the time needed to complete the service at the site in Column $j$. However, if the two sites are considered back-to-back stops, the cost between them is set to 0, indicating a violation. For example, it takes 1439 seconds to travel from stop 50006 to stop 43440 and service stop 43440. Nevertheless, the time distance between stop 43440 and 43448 is 0 since these two stops are in the same location and require the same service, i.e., back-to-back stops.

For easier future understanding for maintenance purposes, the time origin is set to midnight, with the time window adjusted to represent the number of seconds from midnight. Notably, since the earliest time window does not start exactly at midnight, the beginning of the 24-hour cycle does not align with it. As a result, some stops have time windows extending into the next day, and some time window values may exceed 86,400 seconds to indicate this. In other words, the time window can exceed 86,400 seconds (24 hours) if the service window spans into the next day.

Each algorithm discussed in Section 4 is executed for two hours, or a 2-hour run time limit per attempt. This results in four distinct experimental configurations, i.e., Experiments 0-3. To account for possible variability due to random factors, each configuration is tested five times, resulting in a total of 20 independent attempts. We do not use iterations as the criteria for evaluation since different operators often require different run times. Additionally, to better align with customer satisfaction priorities as given by the industrial partner, more emphasis is placed on the feasibility of the solution than the total service time across seven shifts. In other words, a feasible solution is preferred, even if it results in a higher total service time. As a result, Table 2 reports the minimum, maximum, and average total service times across the seven shifts for the final solutions from these five attempts.

**Table 2.** Experimental Results for Different Algorithms Over Five Runs

| Experiment | Algorithm | Algorithm Description | Run Time (Hours) | Minimum Final Total Time (Seconds) | Maximum Final Total Time (Seconds) | Average Final Total Time (Seconds) |
|---|---|---|---|---|---|---|
| 0 | 0 | SSGA | 2 | 308470.0 | 324411.0 | 318559.4 |
| 1 | 1 | ALNS and TA | 2 | 237407.0 | 244912.0 | 240693.2 |
| 2 | 2 | Multiphase ALNS and TA | 2 | 230677.0 | 242304.0 | 236286.4 |
| 3 | 3 | Hybrid multiphase ALNS, TA, and TS | 2 | 231080.0 | 237684.0 | 234734.8 |

As shown in Table 2, the four experiments yield varied performance outcomes. The benchmark experiment (Algorithm 0) shows the worst performance with an average of 324411.0 seconds. Experiment 1 produces the smallest improvements, with an average of 240693.2 seconds, followed by Experiment 2 with an average of 236286.4 seconds. Experiment 3 ranks as the second-best option throughout the study, with an average of 234734.8 seconds. Overall, Algorithm 3, which is used in Experiment 3, produces the best results in Table 2. Noticeably, while the final results across all 15 attempts for Algorithms 1, 2, and 3 are feasible, none of the attempts from Algorithm 0 can find a feasible solution.

To further evaluate the time effect on Algorithm 3 and the effectiveness of TS, Algorithm 3 is run for 10 hours in Experiment 4. Noticeably, similar to Experiment 3, Experiment 4 is also run for 5 times. Table 3 reports the results of both Experiment 3 and Experiment 4. Notably, the results for the 2-hour timeframe presented in Table 3 are those of Experiment 3 as already shown in Table 2.

**Table 3.** Experimental Results on the Effect of Time on Algorithm 3 Over Five Runs

| Experiment | Algorithm | Algorithm Description | Run Time (Hours) | Minimum Final Total Time (Seconds) | Maximum Final Total Time (Seconds) | Average Final Total Time (Seconds) |
|---|---|---|---|---|---|---|
| 3 | 3 | Hybrid multiphase ALNS, TA, and TS | 2 | 231080.0 | 237684.0 | 234734.8 |

| 4 | 3 | Hybrid multiphase ALNS, TA, and TS | 10 | 209567.0 | 225963.0 | 219054.2 |

Experiment 4 delivers a better performance than Experiment 3 by approximately 15000 seconds. Overall, the difference in the average results of the most and the least improved ones (Experiment 4 and Experiment 1, respectively) exceeds 20000 seconds over a 24-hour cycle.

For more illustration, Figure 1 presents convergence graphs to assess the convergence behavior for each experiment, displaying the minimum, maximum, and median values at each iteration. Notably, while different attempts within the same experiment run for the same duration, the number of iterations may vary due to the non-deterministic nature of the roulette wheel. As a result, the later iterations in each convergence graph often contain fewer than five values.

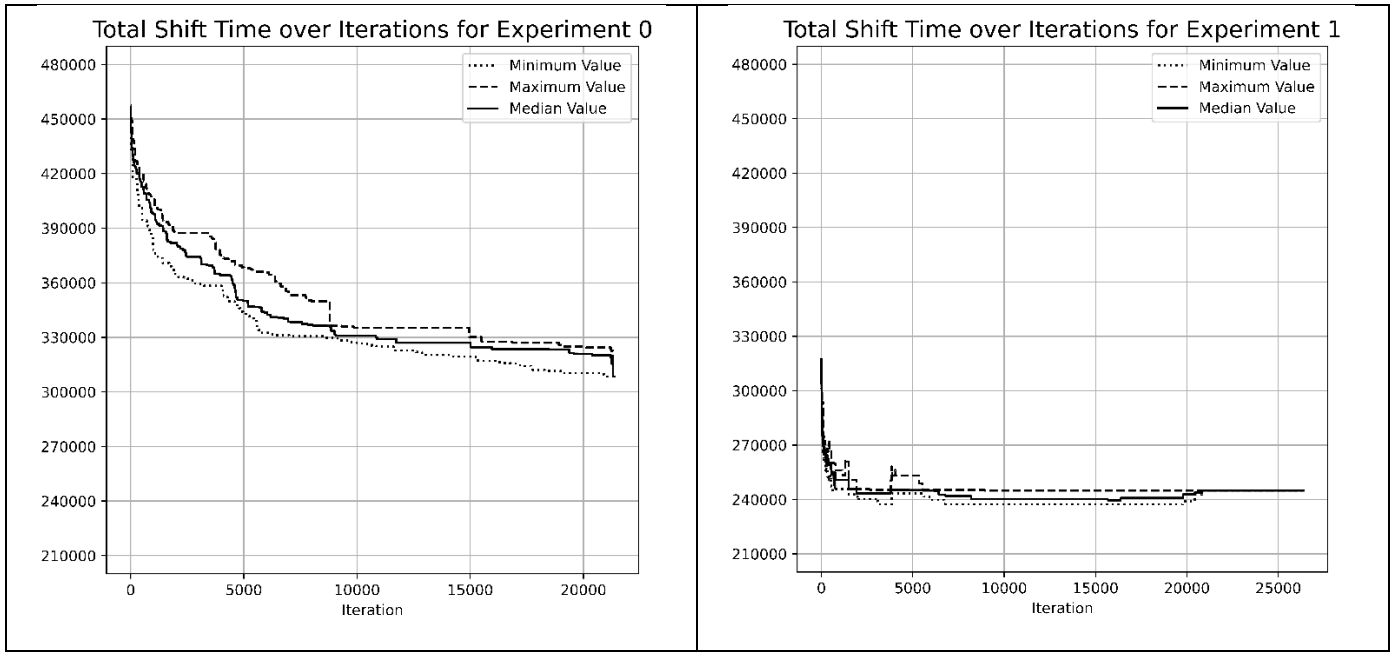

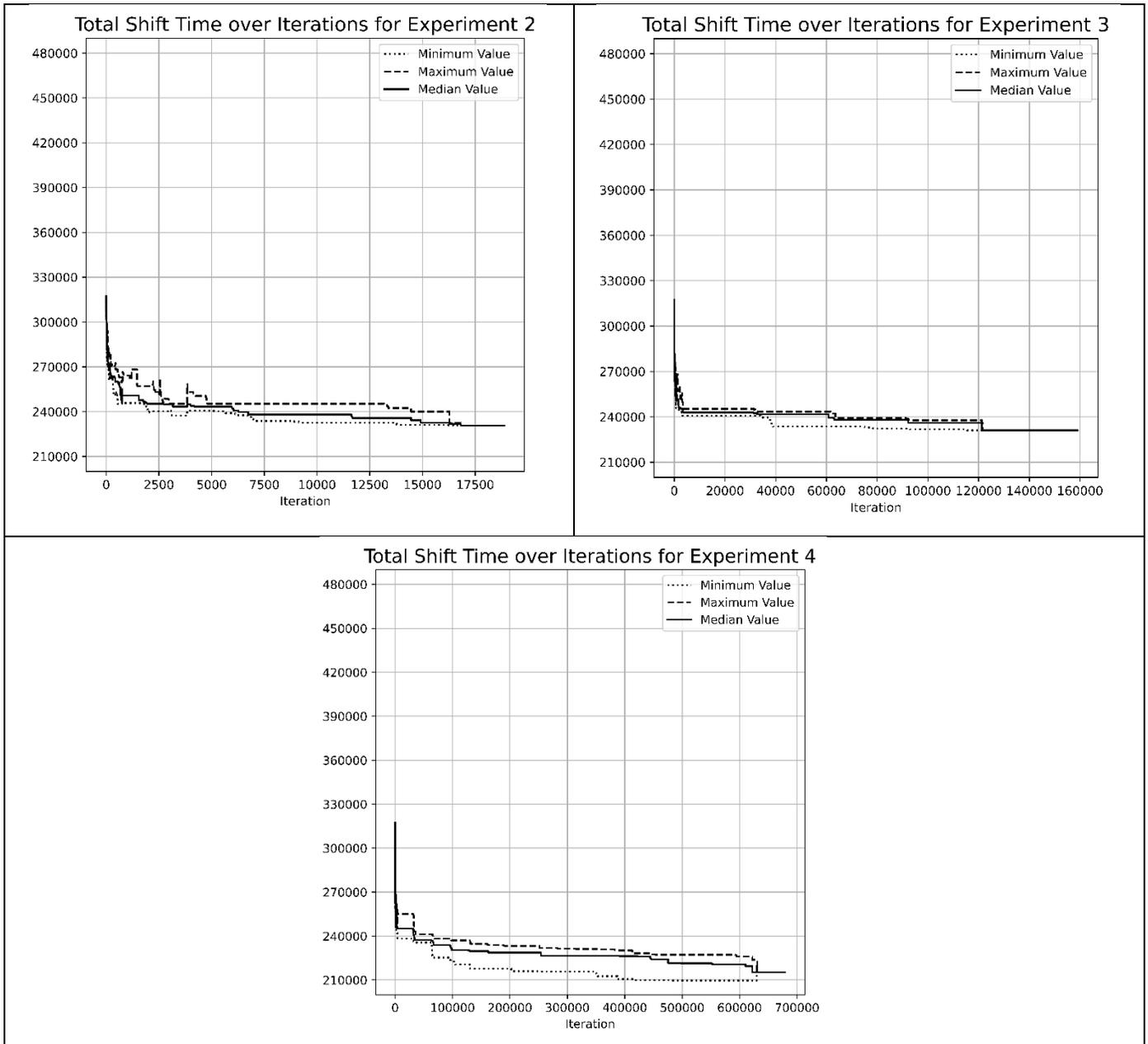

Figure 1. Convergence plots of five different experiments.

All attempts begin with an initial infeasible solution. Hence, all algorithms start by recording the best infeasible solutions, i.e., the solution with the lowest objective function. In the case of Algorithm 0, no feasible solution is recorded since none of the corresponding attempts can identify feasible ones. However, Algorithms 1, 2, and 3 can identify feasible solutions in each experiment. Once feasible solutions are found, these algorithms continue to focus on recording such feasible ones, despite multiple infeasible solutions encountered during the search process.

5.2 Analysis and Discussion

As summarized and illustrated in Tables 2 and 3, although all experiments use the same number of shifts, Experiment 0 shows the worst performance, since the average final total time is the highest and none of the attempts can find a feasible solution. In contrast, the total shift time for Experiment 1, which is adapted based on the example from Wouda and Lan (2023), is approximately 4000 seconds, 6000 seconds, and 20000 seconds higher than that of Experiments 2, 3, and 4, respectively. Notably, Experiments 0, 1, 2, and 3 utilize the same run time (2 hours), while Experiment 4 uses 10 hours. Arguably, this reaffirms the need to develop tailored algorithms to solve VRPSD.

Specifically, notable differences between single-phase and multiphase ALNS are demonstrated in Experiments 1 and 2. The results show that multiphase ALNS improves the overall performance, likely due to its ability to reselect parameters and thresholds, which allows Algorithm 2 to escape local optima to some extent and explore other regions of the solution space. Moreover, a higher number of iterations does not necessarily lead to a better solution, as the algorithm may become trapped in local optima, showing no improvement even with additional time. This explains why the median in Experiment 1 increases toward the end of the graph, as the attempt with the highest number of iterations also yields the worst performance.

Next, differences between multiphase ALNS and the hybrid multiphase ALNS-TS-TA are reflected in Experiments 2, 3, and 4. For clarification, only Experiment 2 uses Algorithm 2, while the remaining two employ Algorithm 3. Furthermore, while Experiments 2 and 3 are run for 2 hours, Experiment 4 is tested for 10 hours. Notably, since one TS iteration is significantly faster than one ALNS iteration generally, the number of iterations in Experiments 3 and 4 is significantly higher than that in Experiments 1 and 2, even when they are run for the same amount of time. Particularly, the performance of Experiment 3 is slightly better with its shorter average time as compared to Experiment 2. This could be because TS enables Algorithm 3 to explore larger regions by accepting all new solutions, thereby moving further away from explored areas. Yet the best result from Experiment 2 is marginally lower than that of Experiment 3.

As discussed in Subsection 5.1, Experiment 4 shows better results than Experiment 3, and both of them outperform Experiments 1 and 2. This reflects the impact of TS and multiphase ALNS implementation which helps the overall Algorithm 3 to escape local optima, a phenomenon already observed in Experiment 1. Evidently, both Experiments 3 and 4 demonstrate small improvements as the time increases. Notably, an approximately 4-hour reduction in the average total travel time is achieved with Experiment 4 versus Experiment 3. This demonstrates that Algorithm 3 can continue to improve the final results if additional run time or resources are applied.

Referring to the two optimization objectives as stated in Section 1, i.e., minimizing the number of shifts and the total travel time duration, Experiments 2, 3, and 4, improve the final results effectively as compared to Experiments 0 and 1. Emphatically, this study evidences positive effects of multiphase ALNS in reducing the total shift time considerably and dealing with strict time window requirements, as shown by differences in results between Experiments 0, 1, and 2. Particularly, such a problem can pragmatically be addressed with TS operations in the hybrid multiphase ALNS-TS-TA algorithm when the running time is properly framed, which is confirmed by the considerable difference of approximately 20000 seconds in the results between Experiment 1 (the least improved) and Experiment 4 (the most improved). Additionally, the variations in results across different attempts of the same experiment in Experiments 1 to 4 demonstrate that Algorithm 1 to Algorithm 3 effectively provided multiple routing solutions, enhancing unpredictability. However, the selected hyperparameters have not been optimized. Therefore, the algorithm could be further improved by choosing the optimal hyperparameters.

## 6. CONCLUSION

This paper deals with Vehicle Routing Problem for Security Dispatch (VRPSD) in the security and patrolling industry which is seemingly less well-known in VRP research-based interest. Specifically, it utilizes the actual input patrolling dataset provided by an industrial partner to realize their optimization assignments, with the technical focus particularly on shift time and route optimization. Accordingly, reflecting stringent requirements observed in security dispatch services, the proposed solution involves an initial greedy algorithm, followed by three main algorithms which utilize different metaheuristics, namely Adaptive Large Neighborhood Search (ALNS), Threshold Accepting (TA), and Tabu Search (TS).

Experimental results demonstrate the value of developing a specific algorithm to solve VRPSD, as shown by the better performance of Experiments 1 to 3 over Experiment 0. Furthermore, out of all tested experiments, Algorithm 3, i.e., the hybrid multiphase ALNS-TS-TA algorithm, significantly reduces the total shift duration. Importantly, the multiphase ALNS outperforms the single-phase ALNS, even when the same computation time is allocated, underscoring the critical role of algorithm design in improving solution quality. As demonstrated by differences in the results of Experiments 3 and 4, the final outcomes can be improved with additional run time or resources. Consequently, the proposed metaheuristic approach, as represented by Algorithm 3, shows high promises to solve VRPSD to meet customers' stringent demands and reduce the total service time and costs.

It is suggested that future research could focus on methods to accelerate convergence, such as integrating Machine Learning techniques to prune non-promising regions of the search space. Additionally, determining automatically optimal hyperparameters for different VRP problems might be a noteworthy research topic. Exploring alternative initialization strategies could also help achieve more stable and faster convergence, improving the quality of the whole algorithm.

## CREDIT AUTHORSHIP CONTRIBUTION STATEMENT

**Nguyen Gia Hien Vu**: Methodology, Writing – original draft, Investigation, Software, Writing – review & editing. **Yifan Tang**: Conceptualization, Validation, Formal analysis. **Rey Lim:** Data curation, Conceptualization. **Gary Wang**: Supervision, Funding Acquisition, Writing – review & editing, Project administration.

## ACKNOWLEDGEMENT

The research grant from MITACS (project number: IT40086) is gratefully acknowledged.

## DATA AVAILABILITY STATEMENT

Detailed input data for this study is available at: https://github.com/vuk1716/VRP-For-Security-Dispatch-Data.

## DECLARATION OF COMPETING INTEREST

The authors declare that they have no known competing financial interests or personal relationships that could have appeared to influence the work reported in this paper.

**APPENDIX 1: NEW OPERATIONS DEVELOPED FOR ALNS**

ALNS serves as a crucial metaheuristic in this study, as it is utilized across Algorithms 1, 2, and 3 outlined in Section 4. In essence, ALNS works by iteratively destroying and repairing parts of the existing solution (Voigt, 2025). During each iteration, ALNS first selects an operation to partially destroy the current solution, then chooses a separate repair operation to restore the destroyed portion, ultimately generating a new solution. The likelihood of selecting a specific operation is determined by its previous performance, predefined initial weights, and a roulette wheel mechanism (Voigt, 2025). The specific destroy and repair operators, which influence the final algorithm's performance (Voigt, 2025), are chosen based on the criteria outlined in Section 3.4.1. For this study, 17 destroy operations are designed and implemented. The detailed description for each operation is as follows:

1. Remove 1 random site from 1 random shift.
2. Remove $n$ random sites from $n$ random shifts. It should be noted that, while the sites must be different, the shifts might appear more than once in the $n$ random shifts.
3. Remove 1 customer that has the lowest service time.
4. Remove $n$ customers that have the lowest service time.
5. Remove 1 random site and its closest neighbor across all shifts. These two sites must not be back-to-back sites.
6. Remove 3 consecutive sites from 1 random shift.
7. Remove $n$ consecutive sites from 1 random shift.
8. Remove 1 customer from the current solution, and consequently the new incomplete solution will have the lowest objective cost if that customer is removed.
9. Remove 1 customer that has the highest lateness. The lateness is defined as the higher between 0 and the difference between the service completion time and the deadline for each site $i$.
10. Remove 2 sites that are the farthest from each other from the same random shift.
11. Remove 2 sites that are farther from each other from the shift that has the highest duration.
12. Remove 1 random customer from the shift that currently has the lowest downtime.
13. Remove the final sites from all shifts that are currently over the shift duration limit (12 hours).
14. Remove the first sites from all shifts that are currently over the shift duration limit (12 hours).
15. Remove 1 customer from the current solution, and consequently the new incomplete solution has the lowest total service time if that customer is removed.
16. Remove 2 sites that are the farthest from each other from the shift with the highest average actual time. The average actual time for a shift is calculated by dividing the current actual service time for the shift by the number of sites serviced along that shift.
17. Remove 2 same-shift sites that are the farthest from each other from all shifts.

Noticeably, multiple operations are designed to focus on late sites, long shifts, or long downtime. For example, Operator 9 is designed to focus on late-serviced customers. There are also multiple operators that include random characteristics.

In addition to destroy operators, 17 repair operators are implemented. The detailed description is described as follows:

1. Insert 1 site into a random shift in a random position. Sites are selected in a random order.
2. Insert 1 site into a random shift in a random position. Sites are selected in the removed order.
3. Insert 1 site into the position that results in the lowest increase in objective cost across all shifts. Sites are selected in a random order.
4. Insert 1 site into the position that results in the lowest increase in objective cost across all shifts. Sites are selected in the removed order.
5. Insert 1 site into the position that results in the lowest increase in actual service time across all shifts. Sites are selected in a random order.
6. Insert 1 site into the position that results in the lowest increase in actual service time across all shifts. Sites are selected in the removed order.
7. Insert 1 site into the position that results in the lowest increase in objective cost across all shifts. Sites with the longest service time are selected first.

8. Insert 1 site into the position that results in the lowest increase in objective cost across all shifts. Sites with the shortest service window are selected first.
9. Insert 1 site into the position that results in the lowest increase in objective cost across all shifts. Sites with the earliest deadline are selected first.
10. Insert 1 site into the position that results in the lowest increase in objective cost across all shifts. Sites that are the closest to the depot are inserted first.
11. Insert 1 site into the position that results in the lowest increase in objective cost across all shifts. Sites that are the farthest to the depot are inserted first.
12. Insert 1 site into the shift with the highest downtime in a random position. Sites are selected in the removed order.
13. Insert 1 site into the shift with the highest downtime into the position resulting in the lowest objective cost increase. Sites are selected in the removed order.
14. Insert 1 site into the shortest shift in a random position. Sites are selected in the removed order.
15. Insert 1 site into the shortest shift in the position resulting in the lowest objective cost increase. Sites are selected in the removed order.
16. Insert 1 site into the position that results in the lowest increase in objective cost across all shifts. Sites are chosen using the regret heuristic.
17. Insert 1 site into the position that results in the lowest increase in actual service time across all shifts. Sites are chosen using the regret heuristic.

Notably, these operations are designed with due consideration of VRPSD characteristics and requirements.